\documentclass{svproc}
\usepackage{algorithm}
\usepackage{algpseudocode}

\usepackage{tikz}
\usepackage{overpic}
\usepackage{booktabs}
\usepackage{amsmath}
\usepackage{amsfonts}
\usepackage{amssymb}
\usepackage{url}
\usepackage{xfrac}

\begin{document}
\mainmatter              
%
\title{Online Performance Estimation with Unlabeled Data: A Bayesian Application of the Hui-Walter Paradigm}
\titlerunning{Online Machine Learning Performance Estimation on Unlabled Data}  
%
\author{Kevin Slote\inst{1} \and Elaine Lee\inst{2}}
\authorrunning{Kevin Slote, Elaine Lee} 
%
\tocauthor{Kevin Slote, Elaine Lee}
\institute{Georgia State University, \\
Department of Mathematics and Statistics, \\
Atlanta GA 30650, USA,\\
\email{kslote1@gsu.edu}
\and
Data Scientist, \\
Hewlett-Packard}

\maketitle              

\begin{abstract}
In the industrial practice of machine learning and statistical modeling, practitioners often work under the assumption of accessible, static, labeled data for evaluation and training. However, this assumption often deviates from reality, where data may be private, encrypted, difficult-to-measure, or unlabeled. In this paper, we bridge this gap by adapting the Hui-Walter paradigm, a method traditionally applied in epidemiology and medicine, to the field of machine learning. This approach enables us to estimate key performance metrics such as false positive rate, false negative rate, and priors in scenarios where no ground truth is available. We further extend this paradigm for handling online data, opening up new possibilities for dynamic data environments. Our methodology involves partitioning data into latent classes to simulate multiple data populations (if natural populations are unavailable) and independently training models to replicate multiple tests. By cross-tabulating binary outcomes across multiple categorizers and multiple populations, we are able to estimate unknown parameters through Gibbs sampling, eliminating the need for ground-truth or labeled data. This paper showcases the potential of our methodology to transform machine learning practices by allowing for accurate model assessment under dynamic and uncertain data conditions.
\end{abstract}
%
%



\section{Introduction}

It is well known that data volume is growing rapidly. Most statistical, machine learning (ML), and deep learning techniques have developed when data were more scarce, and applications make increasingly unverified predictions based on data that is the same or similar to training data. Traditional statistical textbooks go to great lengths to emphasize the difference between interpolation and extrapolation, while warning of the dangers of the latter. Much of these data are delivered online (or streamed) and unlabeled. Researchers trying to move machine learning online still assume that the data are mostly labeled \cite{onlineML,online_convex}. Additionally, performance evaluation is a necessary and often daunting task when using a ML system in the production environment. Performance is often difficult to measure due to insufficient quantities and the variety of ground truth data labeled. Furthermore, data can be confidential, encrypted, or impossible (including costly) to measure, and the data landscape for online or streaming algorithms can constantly change. The actual distribution of the population is often unknown. Thus, it is impossible to relatively easily compare the output distribution of ML systems with the distribution of the population. Fortunately, the problems described are not limited to machine learning. 

Epidemiology is another discipline that relies on accurate efficacy measurements with insufficient ground truth and population data. Disease testing is the most recognizable example of epidemiology. Epidemiologists distinguish between efficacy and effectiveness when talking about moving from a laboratory environment where controlled experiments are possible (efficacy), whereby the test is deployed to test-recipient populations (effectiveness) \cite{Singal2014}. By analogy, a laboratory-developed Covid test measures efficacy by false-negative and false-positive rates in a double-blind controlled trial. The effectiveness is then measured by deploying the test to a remote population. In an applied data science environment, this mimics the applied data scientist who trained a model in a notebook with labeled data and then deployed the model to production.  Thus, the method we propose borrows concepts from how epidemiologists measure efficacy or effectiveness and apply them to measuring the performance of ML systems in production environments where there is no gold standard or ground truth. Our approach can be applied to any field in which unsupervised learning is ubiquitous and is generally likely to be of interest to anyone working in an applied data science environment. Alternatives to our approach are not common.  Based on our research in this area, the most similar approach is to analyze the agreement between multiple classifiers to arrive at an estimate of performance; we provide comparisons of our approach to this strategy.

We use the Hui-Walter paradigm, which uses Bayesian methods to estimate the prior probability (prevalence) and false positive and false negative rates in ensemble models trained to detect the results of running in production or online \cite{Hui1980}. This estimate uses binary results of diagnostic tests, i.e. Covid positive or negative, for multiple tests and multiple populations in a table. We then performed Gibbs sampling \cite{gibbs,gibbs_marginals} to estimate unknown parameters, priors, false positive, and false negative rates without relying on ground truth data. We can make these estimates because more than one population provides us with enough equations and unknowns to solve the parameters of interest. We expand this idea by introducing a closed form solution in which two populations naturally occur, i.e. multiple data centers or regions, and use it to extend this framework to the online environment. When only one population is available, we use latent structure models for both online and offline methods. Many applications use bagging techniques or other ensemble methods in many settings. These ensemble methods naturally extend to our methodology. In our experiments, we divided the Wisconsin breast cancer and Adult data sets into latent classes for demonstration purposes, to simulate scenarios with multiple population data sets, and train independent models for simulation of multiple tests.

\section{Related Work}

The paper ``Estimating Accuracy from Unlabeled Data: a Bayesian Approach" presents methods for estimating classifier accuracies using unlabeled data, focusing on agreement rate-based methods and Bayesian graphical models. Their method relies on agreement rates between classifiers, assuming independent errors, and operates within a static data environment. While effective for static data sets, their framework does not naturally extend to online or dynamic data streams. In contrast, Slote and Lee also assume independent classifiers but adapt the Hui-Walter paradigm from epidemiology, which allows for performance estimation across multiple populations or latent classes. Our work employs a complementary approach that estimates accuracy but focuses on multiple classifiers and multiple populations in an online and static setting, which could offer a practical variation for similar unsupervised learning tasks \cite{Estimating_accuracy_from_unlabeled_data}. Due to the differences in approach, we employ the Rand index (RI) for comparison to our model.

\section{Main results}
\label{sec:main}

The Hui-Walter paradigm is a technique epidemiologists use to measure effectiveness without establishing the ground truth on new data. The paradigm is based on the comparison of multiple tests and their results applied to more than one logical subset of the population \cite{Hui1980}. When evaluating ML systems, tests are analogous to binary categorizers, results are predictions (i.e. a $1$ or $0$ binary classification corresponds to a positive or negative test result), and population subsets are groups of observations that are considered to be in distinct subpopulations. To obtain different subpopulations when only one exists, which may be the case when evaluating some ML systems, we propose using latent structural analysis, consisting of latent class analysis (LCA) or latent profile analysis (LPA), to split the data into two populations. Of course, some ML systems may already use multiple distinct subpopulations; for example, a practitioner has different data centers or databases to choose from.  The framework we propose is two-fold. For machine learning practitioners working in an industry where multiple populations are often available for selection, we may consider the Hui-Walter paradigm applied to two or more models trained to detect results based on these data. If only one population is present, we use LPA or LCA to separate our data into multiple latent populations. Next, we give a closed-form solution applying Hui-Walter to the data stream, measuring prior probabilities and false positive and false negative rates for unlabeled data streams. Our proposed solution dynamically assigns latent classes to streaming data and calculates unknown performance metrics for unlabeled data streams.


\section{Hui-Walter}

The Hui-Walter paradigm suggests that we can form a $n\times m \times k$ contingency table for $n$ binary categorizers, $m$ populations, and $k$ classes. The goal of this method is to estimate a parameter vector $(\hat{\alpha}, \hat{\beta}, \hat{\theta})$ for each model and population, where $\alpha$ is the false positive rate or $1 - \text{specificity}$, $\beta$ is the false negative rate or $1 -\text{sensitivity}$, and $\theta$ is the prior probability of the positive class in the population. The Hui-Walter paradigm is valid for $m$ populations, $n$ classes, and $k$ categorizers where $m,n,k\in \mathbb{N}$. When $m=n=k=2$, we can show this result using maximum likelihood estimation \cite{Hui1980}.

\begin{figure}[H]
\caption{Two tests on one population.}
\label{fig:one-pop}
\centering
\begin{tikzpicture}[scale=2] 
    \draw[thick] (0,0) -- (2.2,0);
    \draw[thick] (0,0) -- (0, 2.2);
    \draw[thick] (2.2,2.2) -- (2.2, 0);
    \draw[thick] (2.2,2.2) -- (0, 2.2);

    \draw[thick] (-0.3, 1.1) -- (2.2, 1.1);
    \draw[thick] (1.1, 0) -- (1.1, 2.5);

    \coordinate[label=left:($+$)] (p1) at (-0.1,1.6);
    \coordinate[label=left:($-$)] (p2) at (-0.1,0.4);

    \coordinate[label=above:($+$)] (p3) at (0.55, 2.2);
    \coordinate[label=above:($-$)] (p4) at (1.65, 2.2);

    \coordinate[label=above:\textbf{$T_2$}] (p5) at (1.1, 2.5);
    \coordinate[label=left:\textbf{$T_1$}] (p6) at (-0.3, 1.1);

    \coordinate[label={\Large $X_1\,$}] (x1) at (0.55, 1.50);
    \coordinate[label={\Large $X_2\,$}] (x2) at (1.65, 1.50);
    \coordinate[label={\Large $X_3\,$}] (x3) at (0.55, 0.40);
    \coordinate[label={\Large $X_4\,$}] (x4) at (1.65, 0.40);
\end{tikzpicture}

\end{figure}

For one population, consider Figure \ref{fig:one-pop}. One population is multinomial distributed cell data. The cells $X_1, X_4$ are where both models ($T_1$ and $T_2$) agree, and $X_2, X_3$ are where the models disagree.

Now, the advantage of maximum likelihood for a single population is that $p_1$ is the probability that both tests produce a positive result. We solve this overdetermined system by minimizing the likelihood function. In our simplified setup, we have a parameter vector $p=(p_1,p_2,p_3,p_4)$, which we wish to estimate. Because these probabilities in question are the outcomes of binary classifiers or diagnostic tests, we may express them in terms of the false positive and false negative rates of the models in question and the population's prior probability. Let $\alpha_1, \alpha_2$ be the false positive rates for $T_1$ and $T_2$, let $\beta_1, \beta_2$ be the false negative rates for $T_1$ and $T_2$, and let $\theta$ be the prior probability for the positive class (base rate) of the population (or the disease in the epidemiological case). 

\begin{equation}
\begin{aligned}
 p_1=& \theta(1-\beta_1)(1-\beta_2) + (1-\theta)\alpha_1\alpha_2 \\
 p_2=& \theta(1 -\beta_1)\beta_2 + (1-\theta)\alpha_1(1-\alpha_2) \\
 p_3=&\theta\beta_1(1-\beta_2) + (1-\theta)(1-\alpha_1)\alpha_2 \\
 p_4=&\theta\beta_1\beta_2 + (1-\theta)(1-\alpha_1)(1-\alpha_2)
\end{aligned}
\label{eq:eq1}
\end{equation}

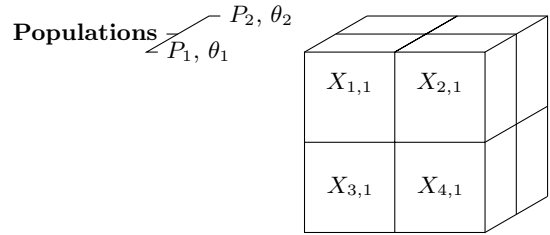
\begin{figure}
    \centering
\newcommand\pgfmathsinandcos[3]{%
  \pgfmathsetmacro#1{sin(#3)} 
  \pgfmathsetmacro#2{cos(#3)}}

\pgfmathsetmacro\angFuite{30}
\pgfmathsetmacro\coeffReduc{0.4}

\begin{tikzpicture}[scale=.3]
\pgfmathsinandcos\sint\cost{\angFuite} %
  \draw (3,0)--(11,0)--(11,8)--(3,8)--(3,0)
        (7,0)--(7,8) (3,4)--(11,4);
          \path[coordinate] (11,4) coordinate (A);

\tikzset{current plane/.estyle={%
  cm={1,0,\coeffReduc*\cost,\coeffReduc*\sint,(0,0)}}}
\begin{scope}[current plane]
\draw  (11,0)--(11,8);
\end{scope}

\tikzset{current plane/.estyle={%
cm={1,0,\coeffReduc*\cost,\coeffReduc*\sint,(0,8)}}}
\begin{scope}[current plane]
   \draw (7,0)--(7,8) (3,4)--(11,4) (7,0)--(7,8);
   \draw (11,0)--(11,8) 
         (3,8)--(3,0) 
         (-3.5,0)--(-4, 0)--(-4,8)--(-3.5,8) node[right] {$P_2$, $\theta_2$};  
          \node[right] at (-3.5,0){$P_1$,  $\theta_1$}; 
          \node[right] at (6,-7){$X_{1,1}$}; 
          \node[right] at (10,-7){$X_{2,1}$};
          \node[right] at (14,-30){$X_{3,1}$}; 
          \node[right] at (18,-30){$X_{4,1}$}; 
   \draw (-4,4) -- (-4.5,4) node [left]{\textbf{Populations}}; 
    \path[coordinate] (11,4) coordinate (C);   
\end{scope} 

\tikzset{current plane/.estyle={%
cm={1,0,0,1,(8*\coeffReduc*\cost,8*\coeffReduc*\sint)}}}
\begin{scope}[current plane]
  \draw (11,0)--(11,8)--(3,8);
  \path[coordinate] (11,4) coordinate (B);
\end{scope}

\draw (A)--(B);
\draw (C)--++(0,-8); 
\end{tikzpicture}
    \caption{Hui-Walter assumptions with a $2 \times 2 \times 2$ contingency table. Variables $X_{1,1}, X_{2,1}, X_{3,1}$ and $X_{4,1}$ are cell frequency counts from a product-multinomial distribution. This setup is the minimum number of populations and classifiers required for the Hui-Walter method. However, this framework supports $n$ populations and $m$ classifiers.}
    \label{fig:two-pop}
\end{figure}

\noindent
We need to consider two populations because the base rates of a single population are unknown. We assume that $\alpha_i$ and $\beta_i$ are the same in both populations for $i=1,2$, therefore we have six equations with six unknowns. In other words, $p_i$ where $i = 1, 2, \dots 8$, can be written algebraically as combinations of 6 variables.  Therefore, we use two populations so that the above estimates are solvable \cite{Johnson2001}. At least two populations guarantee that we have as many equations as unknowns. For two populations, we have two independent multinomial distributions. Therefore, the likelihood function is the product of the likelihood functions \cite{Hui1980}. Now consider two populations, $P_1$ and $P_2$ with prior probabilities $\theta_1$ and $\theta_2$. In that case, our two-by-two-cell data becomes Figure \ref{fig:two-pop}.

\noindent
Now that we have two populations, we can solve the prior probabilities to eliminate the unknowns in our estimates. Because we have two independent multinomial distributions, which can be considered a product-multinomial distribution, the likelihood function is the product of the individual likelihood functions adjusted with the sample size. Let $n_1, n_2$ be the total sample size of each population, then the likelihood function is 


$$l = l_1l_2 =\prod_j^2 \prod_i^2 p_{i,j}^{x_{ij}}$$

\noindent
Where $p_{i,j}$ is the probability of observing the $j$ th outcome in the $i$ th population.

We want to estimate the parameter vector $\boldsymbol{X}=(\mathbf{\alpha}, \mathbf{\beta}, \mathbf{\theta})$. A real solution with coordinates that satisfy these constraints is a maximum likelihood estimate. 

\noindent
For our simplified case with two populations we get 
$$\frac{\partial L}{\partial p_i} = \frac{x_i}{p_i} - \lambda = 0,\frac{\partial L}{\partial p_j} =  \frac{x_j}{p_j} - \mu = 0$$

\noindent
for $i=1,\dots,4$, and for $j=5,\dots,8$. The maximum likelihood then is $\mathbf{p}=(\frac{p_1}{n_1}, \dots, \frac{p_4}{n_1}, \frac{p_5}{n_2}, \dots, \frac{p_8}{n_2})$ from equation \ref{eq:eq1}. This closed formula gives the maximum likelihood estimate for the product-multinomial distribution. This MLE extends naturally to setups with more than 2 classes.

In other cases, when the points of the global maximum are complex or lie outside the unit hypercube, estimates can be obtained by numerical maximization of $l$ with the solution restricted to be within the unit hypercube, again subject to appropriate constraints. Asymptotically, because the observed proportions of cell counts are continuous functions of the parameters, the maximum likelihood solution will converge to $(\mathbf{\alpha}, \mathbf{\beta}, \mathbf{\theta})$. Let $L=ln(l)$, then we can formulate the information matrix 
$$\mathcal{I}(\boldsymbol{X}) =  \mathbb{E} \left[ \operatorname{Hess} (\boldsymbol{X} \otimes \boldsymbol{X}) \right] \text{.}$$

\noindent
In other words, when the MLE resides within the interior point of the feasible set, we can solve for the information matrix by taking the expectation value of the hessian of the outer product of our parameter vector, $\boldsymbol{X}$ \cite{Blasques2018}. 

Inverting this matrix gives the variance and covariance matrix. Now that we have shown that there is a maximum likelihood estimate for the parameter vector $(\mathbf{\alpha}, \mathbf{\beta}, \mathbf{\theta})$ exists for the $2\times 2 \times 2$ case, it is trivial to see that this generalizes to the $n\times m \times k$ case for arbitrary dimensions of the population outcome tensor. We can generally find a solution to the parameter vector using Bayesian methods such as Monte Carlo estimates or Gibbs sampling to estimate our parameter vector, $\boldsymbol{X}=(\mathbf{\alpha}, \mathbf{\beta}, \mathbf{\theta})$.

One of the assumptions of Hui-Walter is that the tests (or ML binary classifiers) be conditionally independent \cite{Hui1980}. We can test for conditional independence, which can easily be tested by performing a goodness-of-fit hypothesis test on the contingency table with a $G$-test \cite{hui2}.

\section{Hui-Walter Online}

Online Machine Learning refers to training machine learning models on data that arrive continuously and in real time rather than using a static data set. In this setting, the model updates its parameters as new data arrive and makes predictions accordingly \cite{onlineML}. This approach is helpful in applications such as streaming data analysis, recommendation systems, and dynamic pricing.

One of the key benefits of online machine learning is its ability to handle large amounts of data efficiently, which is especially important in scenarios where data volume is increasing rapidly. Additionally, online machine learning algorithms can adapt to changing data distributions, which is critical in applications where the underlying data distribution constantly evolves. This process enables the models to provide accurate predictions even in dynamic environments, improving the overall system performance.

Various algorithms work online, including stochastic gradient descent \cite{benarous2024online}, k-means \cite{pmlr_online_k_means}, linear regression \cite{online_linear_regression}, and online support vector machines \cite{onlineML}. These algorithms differ in their update rules, but all aim to minimize the cumulative loss incurred over time as new data arrive. A critical aspect of online machine learning is the choice of the loss function, which determines the trade-off between the accuracy of the model and its ability to adapt to new data. However, these methods are still limited to using labeled data and assuming that ground truth is available. Online machine learning provides a powerful and flexible approach to real-time training machine learning models, but fails in some applied settings where the data is encrypted. 

We saw previously that the maximum likelihood estimate for the product-multinomial distribution is $\mathbf{p}=(\frac{p_1}{n_1}, \dots, \frac{p_4}{n_1}, \frac{p_5}{n_2}, \dots, \frac{p_8}{n_2})$. We use this estimate to come up with a naive estimate of the base rate for a population taking $\frac{X_i}{n}$ for $i=1,2$. We use this estimate to solve for the other parameters related to false-positive and false-negative rates and keep a tally of the streaming data as a time series. We then use the prior probability of our base class from our training environment to test for significant prior drift.

A closed-form solution to the Hui-Walter estimates without a confidence interval appears in \cite{Hui1980,enola2000}. We reformulate our contingency table in tabel \ref{tab:exp} as follows:

\begin{table}[!htb]
\centering
\begin{tabular}{|l|l|l|l|}
\hline
            & \begin{tabular}[c]{@{}l@{}}$T_2$\\ (pos)\end{tabular} & \begin{tabular}[c]{@{}l@{}}$T_2$\\ (neg)\end{tabular} &       \\ \hline
$T_1$ (pos) & $a_i$                                                 & $b_i$                                                 & $g_i$ \\ \hline
$T_1$ (neg) & $c_i$                                                 & $d_i$                                                 & $h_i$ \\ \hline
            & $e_i$                                                 & $f_i$                                                 & $n_i$ \\ \hline
\end{tabular}
\vspace{5mm}
\caption{Explanation of variables in terms of model agreement and disagreement indexed by population.}
\label{tab:exp}
\end{table}

\noindent
where $i$ ranges over each population. For simplicity, we will focus on $i=2$ but mention that this can be extended to more populations and classes. By computing the row sums and column sums of this matrix, we look at the following discriminant:

\begin{equation}
F=\pm \sqrt{\left(\frac{g_1e_2 - g_2 e_1}{n_1n_2} + \frac{a_1}{n_1} - \frac{a_2}{n_2}\right)^2  -4 \left(\frac{g_1}{n_1} - \frac{g_2}{n_2}\right)\frac{a_1e_2 - a_2 e_1}{n_1n_2 }}
\label{eq:eq2}
\end{equation}

\noindent
If this discriminant is zero or complex, then this online method cannot be calculated \cite{Hui1980}. One possible explanation for this phenomenon is the existence of Simpson's paradox and the algebraic geometry that arises when dealing with three-way contingency tables \cite{algebra}. The closed solutions to the Hui-Walter estimates are as follows: 

\begin{equation}
\begin{aligned}
    \hat{\theta_1} &= \frac{1}{2} -  \left[ \frac{g_1}{n_1}\left(\frac{e_1}{n_1} - \frac{e_2}{n_2}\right) + \frac{g_1}{n_1}\left(\frac{g_1}{n_1} - \frac{g_2}{n_2}\right) + \frac{a_2}{n_2} - \frac{a_1}{n_1} \right]\frac{1}{2F} \\
    \hat{\theta_2} &= \frac{1}{2} -  \left[ \frac{g_2}{n_2}\left(\frac{e_1}{n_1} - \frac{e_2}{n_2}\right) + \frac{g_2}{n_2}\left(\frac{g_1}{n_1} - \frac{g_2}{n_2}\right) + \frac{a_2}{n_2} - \frac{a_1}{n_1} \right] \frac{1}{2F} \\
    \hat{\alpha_1} &= \left(\frac{g_1e_2 - e_1g_2}{n_1n_2} + \frac{a_2}{n_2} - \frac{a_1}{n_1} + F \right)\left[2\left(\frac{e_2}{n_2} - \frac{e_1}{n_1}\right)\right]^{-1} \\
    \hat{\alpha_2} &= \left(\frac{g_2e_1 - e_2g_1}{n_1n_2} + \frac{a_2}{n_2} - \frac{a_1}{n_1} + F \right)\left[2\left(\frac{g_2}{n_2} - \frac{g_1}{n_1}\right)\right]^{-1}  \\
    \hat{\beta_1 } &= \left(\frac{f_1h_2 - h_1f_2}{n_1n_2} + \frac{d_2}{n_2} - \frac{d_1}{n_1} + F \right)\left[2\left(\frac{e_2}{n_2} - \frac{e_1}{n_1}\right)\right]^{-1} \\
    \hat{\beta_2 } &= \left(\frac{f_2h_1 - h_2f_1}{n_1n_2} + \frac{d_2}{n_2} - \frac{d_1}{n_1} + F \right)\left[2\left(\frac{g_2}{n_2} - \frac{g_1}{n_1}\right)\right]^{-1}
\end{aligned}
\label{eq:eq3}
\end{equation}

\noindent
Using the formulas mentioned above, we can increase the streams' counts and apply the estimates to the streaming data. This allows us to investigate the false-positive and false-negative rates, as well as the prior probabilities of streaming data at a specific time step. Interestingly, boosting provides an approach to this problem. It is a family of algorithms that convert multiple weak learners working in parallel into strong learners. This is where Hui-Walter online comes into play. With it, we can harness the power of these weak learners to assess the false-positive and false-negative rates, enhancing predictions by making them ``prior aware" of the unlabeled data streams. This approach is not unique. In fact, several learning frameworks, such as the one mentioned in \cite{Davis2020}, advocate better predictions by making the learner aware of previous information. However, a point to be taken into account is the sign of $F$ from equation \ref{eq:eq2}. As noted in \cite{Hui1980,Johnson2001}, it can be positive or negative, and sometimes it does not present a solution at all. Hence, its value is determined by what yields \textit{plausible} solutions, specifically within a probability simplex.

\noindent
Our online algorithm runs as follows in Algorithm \ref{alg:online_hw}:

\begin{algorithm}[H]
\caption{Online Hui-Walter on Streaming Data}
\label{alg:online_hw}
\begin{algorithmic}
\Procedure{Online Hui-Walter}{$\mathcal{X}, T$} 
 \State $t \gets 1$ 
 \State Initialize models $f_1, f_2$ with $\mathcal{X}$ 
 \State Initialize three-way contingency table $\mathcal{T}$ and history $\mathcal{H}$ 
    \While{$t \leq T$}
        \State Receive instance $x_t$ 
        \State Predict labels $y_1 = f_1(x_t), y_2 = f_2(x_t)$
        \State Add $x_t$ to $\mathcal{H}$
        \State Get LPA profiles of $\mathcal{H}$
        \State Update cell counts of $\mathcal{T}$
        \If {t $>>$ n}
            \State Calculate $\mathbf{X}=(\mathbf{\theta}, \mathbf{\alpha}, \mathbf{\beta}))$
        \EndIf
        \State $t \gets t + 1$
    \EndWhile
\EndProcedure
\end{algorithmic}
\caption{Online algorithm to calculate model performance.}
\end{algorithm}

In Algorithm \ref{alg:online_hw}, $\mathcal{X}$ is the data set and $f_1$ and $f_2$ trained on subsets of columns of $\mathcal{X}$, $T$ is the number of instances to receive, $\mathcal{T}$ is a contingency table with as many dimensions as populations, $\mathcal{H}$ is the history of previously seen samples. 
After the $n$ instances have been predicted, the estimates for $\theta, \alpha, \beta$ start to be calculated and updated.  The selection of $n$ should be reasonable such that there is enough stability in the values for $\theta, \alpha, \beta$ (lessening variation from one instance to the next) without sacrificing too many of the samples.  This analogous to a ``burn-in" concept commonly seen in MCMC \cite{burnin}, but keep in mind that we are not discarding samples or assuming stationarity has been achieved.

\section{Data Sets}
\label{sec:experiments}

In this study, we used two canonical data sets, the Wisconsin Breast Cancer data set and the Adult data set, both commonly employed in machine learning research for benchmarking and classification tasks. We selected these data sets because of their well-documented attributes and applicability to evaluate the performance of classifiers. In the following, we provide an overview of each data set and the specific features used in our analysis.


\subsection{Wisconsin Breast Cancer}
The Wisconsin breast cancer data set is a clinical data collection used to analyze breast cancer tumors. It was created in the early 1990s by Dr. William H. Wolberg, a physician and researcher at the University of Wisconsin-Madison. The data set contains 569 observations and includes information on patient characteristics, such as age, menopausal status, and tumor size, as well as diagnosis and treatment of cancer.  In the diagnosis of 569 observations, 357 were benign and 212 were malicious.

The Wisconsin Breast Cancer Data Set has been widely used in machine learning and data mining to predict breast cancer diagnosis and prognosis. Researchers have used the data set to develop algorithms and models to classify tumors as benign or malignant and predict the probability of breast cancer recurrence or survival. These predictions can guide the treatment of patients with breast cancer and identify potential risk factors for the disease.

In addition to its role in medicine, it has been used as an almost canonical data set for data science education and benchmarking of ML algorithms. It is one of the most downloaded data sets on the UCI Machine Learning Repository\cite{uci}\footnote{https://archive.ics.uci.edu/ml/datasets/Breast+Cancer+Wisconsin+\%28Diagnostic\%29}.

Our adoption of a hierarchical clustering-based feature selection strategy \cite{heirarchal_feautre_selection} was a significant step in our efforts to reduce dimensionality and tackle multicollinearity in our data set. This approach, designed to eliminate redundant information and simplify the model, not only enhances interpretability and computational efficiency, but also plays a crucial role in preventing overfitting and improving the generalization performance of the model on unseen data \cite{mulit_colin}.

We started by calculating the Spearman rank-order correlation matrix \cite{spearman} for all characteristics. Spearman correlation effectively captures monotonic relationships between variables, accommodating linear and non-linear associations. This advantage makes it suitable for identifying linear and non-linear feature dependencies.

To prepare the data for clustering, we transformed the correlation matrix into a distance matrix \cite{distance_clustering} using the formula $D_{ij} = 1 - |\rho_{ij}|$, where $\rho_{ij}$ is the absolute Spearman correlation coefficient between features $i$ and $j$. This transformation ensures that highly correlated features—positively or negatively correlated—have smaller distances between them. As a result, features that share similar information are positioned closer together in the distance space.

Our use of Ward's linkage method \cite{wald} in the hierarchical clustering process was instrumental in promoting the formation of compact and well-separated clusters. This method, by minimizing the total within-cluster variance, effectively identifies natural groupings in the data, where features within the same cluster are highly similar, and those in different clusters are dissimilar.

We formed clusters of the data by applying a distance threshold of $0.9$ to the hierarchical clustering dendrogram. This threshold grouped highly correlated features, effectively partitioning the set of characteristics into distinct clusters based on similarity. By setting this threshold, we controlled the granularity of the clustering, balancing retaining important features and reducing redundancy.

We selected one representative feature from each cluster \cite{feature_selection_from_clusters} to include in our reduced feature set. The selection criterion was straightforward: we chose the first feature in each group. This approach simplifies the selection process, though alternative criteria could also be applied, such as selecting the feature with the highest variance or domain-specific importance. Following this procedure, we reduce the number of features from the original set to a smaller subset of representative features.

This approach selected five representative features: Mean Radius, Mean Texture, Mean Smoothness, Mean Compactness, and Texture Error (Texture SE). For each selected feature, we calculated the average intra-cluster and inter-cluster correlations to assess the effectiveness of our clustering method. The Mean Radius had an average intra-cluster correlation of $0.848 \pm 0.186$, significantly higher than its average inter-cluster correlation of $0.338 \pm 0.213$ ($p = 2.237e-05$). Similarly, Mean Texture showed an average intra-cluster correlation of $0.909 \pm 0.000$ versus an inter-cluster correlation of $0.254 \pm 0.129$ ($p = 3.448e-02)$. The other selected features also demonstrated higher intra-cluster correlations compared to inter-cluster correlations, with p-values indicating statistical significance: Mean Smoothness (intra: $0.541 \pm 0.133$, inter: $0.308 \pm 0.171$, $p = 3.185e-03$), Mean Compactness (intra: $0.829 \pm 0.053$, inter: $0.461 \pm 0.183$, $p = 2.330e-07$), and Texture SE (intra: $0.416 \pm 0.027$, inter: $0.169 \pm 0.119$, $p = 2.217e-02$) \footnote{https://github.com/kslote1/hui-walter/blob/main/features.ipynb}.

The cophenetic correlation coefficient of hierarchical clustering, which measures the degree of faithfulness the dendrogram maintains pariwise distances\cite{cophenetic,saracli2013comparison}, was $0.672$, indicating a moderate level of fidelity in representing pairwise feature dissimilarities. We then evaluated the performance of a Random Forest classifier using both the complete feature set and the reduced set of selected features. The model trained on all features achieved an accuracy of $95.6\% \pm 2.3\%$, while the model using the selected features achieved an accuracy of $92.3\% \pm 2.8\%$. The difference in model accuracy was not statistically significant ($p = 0.065$), suggesting that the reduced feature set maintains comparable predictive power. These results demonstrate that our feature selection method effectively reduces dimensionality by eliminating redundant features while preserving the essential information necessary for accurate classification.

\subsection{Adult}
This is another canonical data set from the UCI Machine Learning Repository\cite{uci} \footnote{https://archive.ics.uci.edu/ml/datasets/Adult}.  It contains census data from 1994 for the task of predicting whether an adult's income exceeds \$50k/year.  There are 32,561 observations described with a mix of categorical and continuous characteristics such as age, sex, education level, occupation, capital gains, and capital losses.

\section{Hui-Walter Data Experiments}

The core findings of our experimental results are that for each experimental data set in the static context, we found the true parameters of interest inside the confidence intervals produced by Gibbs Sampling.

\setcounter{table}{0}

\subsection{Wisconsin Breast Cancer Data Set}


\begin{figure}[!htb]
 \centering
  \includegraphics[scale=0.3]{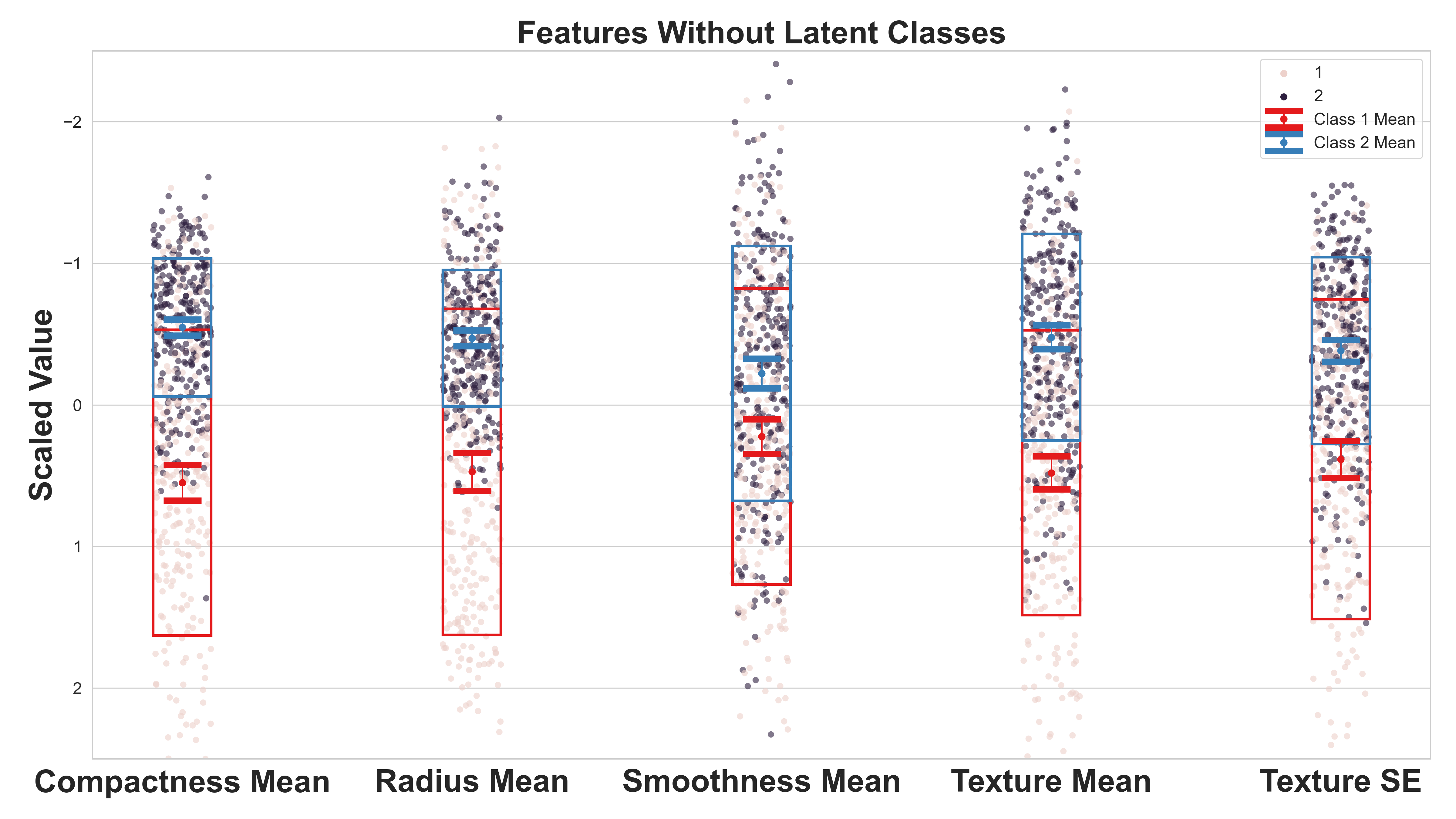}
\vspace{0.5cm} 
  \includegraphics[scale=0.3]{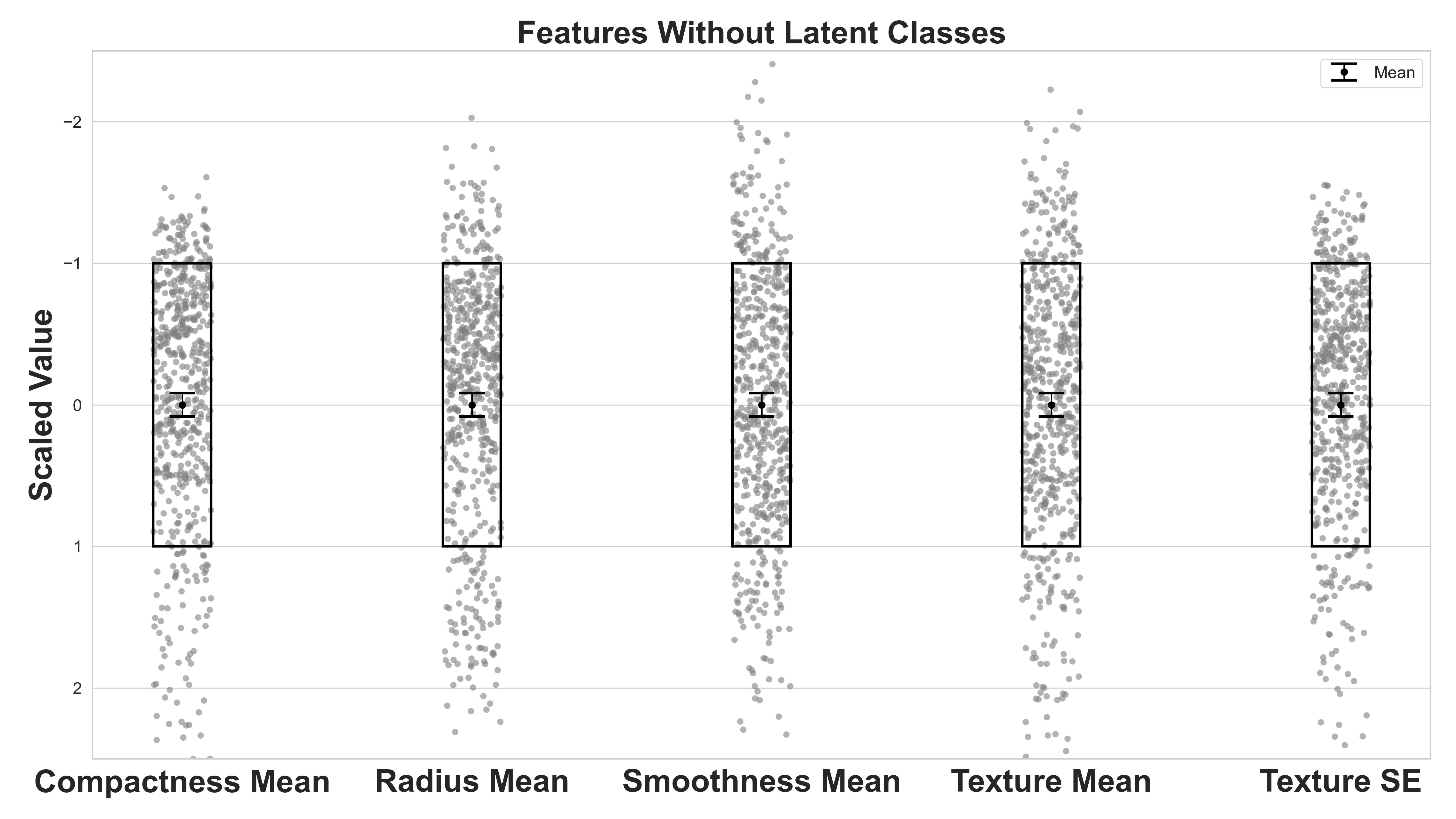}
  \caption{Comparison of scaled features from the Wisconsin Breast Cancer Data Set distributions with and without latent classes. The top panel displays the distributions of five scaled features—\textbf {Compactness Mean}, \textbf {Radius Mean}, \textbf {Smoothness Mean}, \textbf {Texture Mean}, and \textbf {Texture SE}—segmented by latent profile classes. Individual data points are color-coded by class, with means, standard deviations, and 95\% confidence intervals indicated. The bottom panel shows the same features aggregated without class separation, illustrating the overall distributions of the Latent Classes.}
  \label{fig1}
\end{figure}

\begin{figure}[!htb]
  \centering
  \includegraphics[scale=0.7]{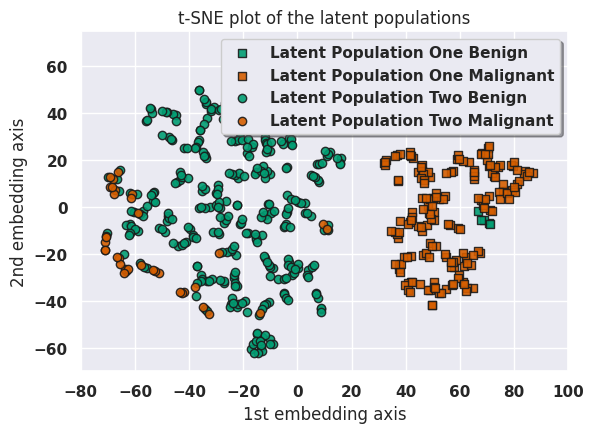}
  \caption{The two-dimensional t-SNE visualization of the training data from the Wisconsin breast cancer data set shows the viability of reducing the data space into Latent Classes. The high-dimensional feature space reduces to two dimensions using t-distributed Stochastic Neighbor Emulation (t-SNE)\cite{tsne} with two components. Each point represents a single patient sample: benign cases are depicted as green squares and malignant cases are represented as red circles. This embedding reveals distinct clustering of benign and malignant samples, indicating that the selected features capture intrinsic differences between the two classes and suggest potential separability in the reduced dimensional space.}
  \label{fig2}
\end{figure}

We performed a Latent Profile Analysis (LPA) on the Wisconsin Breast Cancer data set to obtain two populations. Due to collinearities in the data set, a subset of features was selected to estimate the latent profiles. This subset was determined via hierarchical clustering on Spearman rank-order correlations and selecting one feature from each cluster, where clusters were defined as being separated by at least a distance of 1 per Ward's linkage. LPA, assuming varying variances and varying covariances, was used to fit models assuming 1 through 10 latent classes. The approximate weight of evidence (AWE) criterion was used to identify the optimal number of classes, which was 2. AWE was used instead of BIC for model selection due to its ability to select more parsimonious models \cite{awe}, especially relevant because we assumed complex parameterization (varying variances and varying covariances).

\begin{figure}[!htb]
  \centering

    \includegraphics[width=1.0\textwidth]{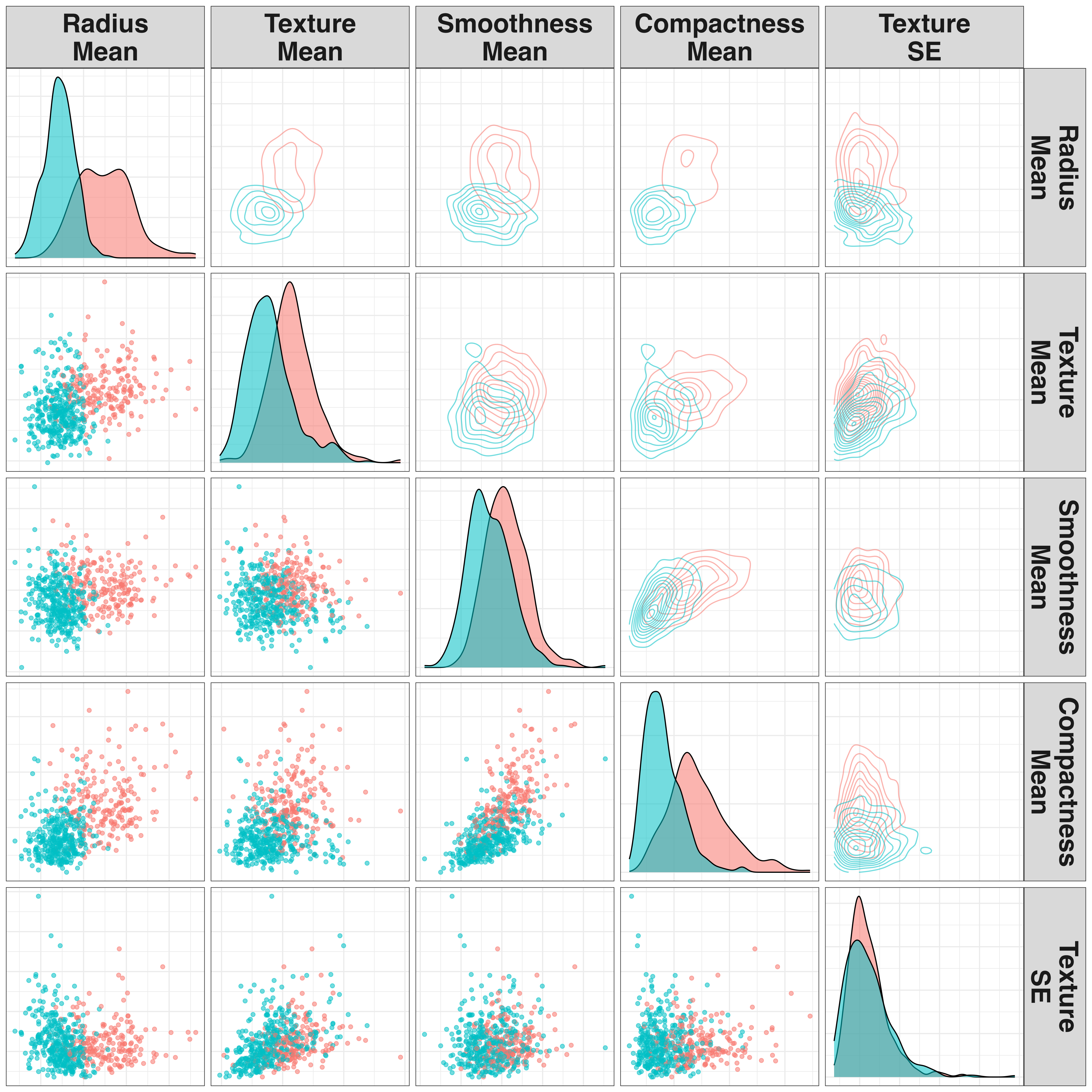}
    \caption{A pairwise scatter plot matrix of selected morphological features from the Wisconsin Breast data set demonstrates the effectiveness of a two-class Latent Class Analysis (LCA) model. Plotted features are \textbf{Radius Mean}, \textbf{Texture Mean}, \textbf{Smoothness Mean}, \textbf{Compactness Mean}, and \textbf{Texture SE}, with data points colored by their assigned latent class. The distinct groupings in the plots indicate that the LCA effectively captures latent patterns in the data.}
  \label{fig3}
\end{figure}

In Figure \ref{fig3}, to identify Latent Classes among the features, a pairwise scatter plot matrix of selected morphological features from the Wisconsin Breast shows the viability of identifying latent classes. The variables plotted are \textit{Radius Mean}, \textit{Texture Mean}, \textit{Smoothness Mean}, \textit{Compactness Mean}, and \textit{Texture SE}. Each distribution represents a tumor feature, with colors indicating the assigned latent class (\textit{Latent Class 1} or \textit{Latent Class 2}) from a two-class LCA model, a critical step in the process performed on binned categories of these variables. The diagonal panels display the density distributions of each feature, highlighting their univariate characteristics. The lower triangle panels present scatter plots between pairs of variables, illustrating bivariate relationships. The upper triangle panels show density plots of these feature pairs, providing additional insights into their joint distributions. Coloring by latent class reveals distinct groupings within the feature space, suggesting that the LCA effectively captures latent patterns in the data. Figure \ref{fig1} shows the columns of the data partitioned into one and then two latent classes. In Figure \ref{fig2}, we performed the t-SNE \cite{tsne} dimension reduction of the columns of the Wisconsin Breast Cancer data set and labeled it according to the latent classes found with the analysis of the latent profile. We see that latent profile analysis and t-SNE separated the data into two latent populations.

We trained a random forest model (RF) and a support vector classifier (SVC) model using the data after the two latent populations have been assigned. To ensure that the models are independent, we chose mutually exclusive columns of the data for each model.  Then, we split the data set into a training and test data set, sizes 200 and 369, respectively.  Finally, we trained the RF and SVC models on the training data set.

\begin{table}[!htb]
\centering
\begin{tabular}{@{}lcrr@{}} 
\toprule 
&  & \multicolumn{2}{c}{Model Outcome}\\\cmidrule{3-4} 
Population & Model & Malignant & Benign\\ \midrule 
Population One  & RF & 109 & 80\\ 
                & SVC & 89 & 100\\ 
Population Two & RF & 10 & 170\\ 
              & SVC & 19 & 161\\\bottomrule 
\end{tabular}
\vspace{5mm}
\caption{Predictions from the model over the two latent populations.}
\label{table2_wbcd}
\end{table}

Table \ref{table2_wbcd} contains a three-way contingency table of categorizations on the Wisconsin Breast Cancer data set within each latent population based on the RF and SVC models. Both models predicted different proportions of malignant to benign tumors for the populations, suggesting that the models have some degree of independence.

With the models trained, we then proceeded to estimate parameters $\hat{\alpha}$, $\hat{\beta}$, and $\hat{\theta}$ without ground truth labels using the Hui-Walter framework with Gibbs sampling.  Sensitivity and specificity were instantiated with a distribution of $Beta (1,1)$.

\begin{table}[!htb]
  \centering
  \begin{tabular}{llllllll}
    \toprule
    \multicolumn{2}{c}{Experimental Results} \\
    \cmidrule(r){1-2}
    Model & TPR & TNR & FPR ($\hat{\alpha}$) & FNR ($\hat{\beta}$) \\
    \midrule
    RF & 0.690 & 0.956 & 0.044 & 0.310\\
    SVC & 0.547 & 0.899 & 0.101 & 0.453\\

    \cmidrule(r){1-2}
    Population & Prior ($\hat{\theta}$) \\
    \midrule
    Population One  & 0.819\\
    Population Two  & 0.035\\

    \bottomrule
  \end{tabular}
  
  \begin{tabular}{llllll}
    \toprule
    \multicolumn{2}{c}{True Values} \\
    \cmidrule(r){1-2}
    Model & TPR & TNR & FPR & FNR \\
    \midrule
    RF & 0.789 & 0.969 & 0.031 & 0.211\\
    SVC & 0.549 & 0.868 & 0.132 & 0.451\\

    \cmidrule(r){1-2}
    Population & Prior\\
    \midrule
    Population One  & 0.635\\
    Population Two  & 0.122\\
    \bottomrule
  \end{tabular}
  \vspace{5mm}
  \caption{Top are the estimated parameters (TPR, TNR, prior) for both models. Below
are the known values from the training environment in the Wisonsin Breast Cancer data Set.}
    \label{table1_wbcd}
\end{table}
The estimated parameters using Hui-Walter with Gibbs Sampling are shown along with the actual values based on the ground truth in Table \ref{table1_wbcd}.  For the SVC model, Hui-Walter estimated $\hat{\alpha}$ within 0.03 (0.101 compared to 0.132) and $\hat{\beta}$ within 0.002 (0.453 compared to 0.451).  For the RF model, the Hui-Walter estimates for $\hat{\alpha}$ were within 0.02 (0.044 compared to 0.031) and $\hat{\beta}$ was within 0.1 (0.310 compared to 0.211).  The estimated priors $\hat{\theta}$ for Populations One and Two were close as well; Hui-Walter overestimated the prevalence in Population One (0.819 compared to 0.635) and underestimated for Population Two (0.035 compared to 0.122).

In Table \ref{table3_wbcd}, we have the confidence intervals for each parameter estimate. All confidence intervals for the errors contain the parameter we found during the evaluation phase of the experiment. However, the true prevalence (prior or base rate) is outside the confidence interval for both populations ($\hat{\theta}_1$ and $\hat{\theta}_2$).  Depending on context, these margins of error may be acceptable as a measure of classifier performance without requiring ground truth labeled data to compute.

\begin{table}[!htb]
  \centering
  \begin{tabular}{llllll}
    \toprule
    Parameter & CI & Mean & SD \\
    \midrule
    $\hat{\theta}_1$ & (0.693, 0.947) & 0.819 & 0.064\\ 
    $\hat{\theta}_2$  & (3.9e-07, 0.084), & 0.035 & 0.026 \\
    $\hat{\alpha}_1$ & (0.004, 0.081) & 0.044 & 0.020 \\ 
    $\hat{\alpha}_2$  & (0.054, 0.150) & 0.101 & 0.025 \\ 
    $\hat{\beta}_1$  & (0.198, 0.425) & 0.310 & 0.058 \\ 
    $\hat{\beta}_2$  & (0.357, 0.547) & 0.453 & 0.049 \\
    \bottomrule
  \end{tabular}
  \vspace{5mm}
  \caption{The estimated parameters and standard deviations from the Hui-Walter
paradigm on static data for the Wisconsin Breast Cancer data set.}
  \label{table3_wbcd}
\end{table}

\subsection{Adult}

In lieu of latent profile analysis, we used the sex of the adult to partition the data set, resulting in two populations of $10,771$ females and $21,790$ males.  We performed an $\sfrac{80}{20}$ split on the Adult data set to produce a training and test data set, sizes $26,048$ and $6,513$, respectively.  We trained two classifiers, a logistic regression (LR) and a random forest classifier (RF), on different columns of the data in the training data set.

\begin{table}[!htb]
\centering
\begin{tabular}{@{}lcrr@{}} 
\toprule 
&  & \multicolumn{2}{c}{Model Outcome}\\\cmidrule{3-4} 
Population & Model & \$50k \& Under & Over \$50k\\ \midrule 

Female  & LR & 2003 & 123\\ 
                & RF & 1968 & 158\\ 
Male & LR & 3521 & 866\\ 
              & RF & 3258 & 1129 \\
\bottomrule 
\end{tabular}
\vspace{5mm}
\caption{Model Predictions Over The Two Subpopulations.}
\label{table2_adult}
\end{table}

Table \ref{table2_adult} contains a three-way contingency table of categorizations on the Adult data set within each  population based on the LR and RF models. Both models predicted similar proportions of income for the populations.

With the models trained, we then proceeded to estimate parameters $\hat{\alpha}$, $\hat{\beta}$, and $\hat{\theta}$ without ground truth labels using the Hui-Walter framework with Gibbs sampling.  Sensitivity and specificity were instantiated with a distribution of $Beta (1,1)$.

\begin{table}[!htb]
  \centering
  \begin{tabular}{llllllll}
    \toprule
    \multicolumn{2}{c}{Experimental Results} \\
    \cmidrule(r){1-2}
    Model & TPR & TNR & FPR ($\hat{\alpha}$) & FNR ($\hat{\beta}$) \\
    \midrule
    LR & 0.608 & 0.992 & 0.008 & 0.392 \\
    RF & 0.793 & 0.990 & 0.010 & 0.207 \\

    \cmidrule(r){1-2}
    Population & Prior ($\hat{\theta}$) \\
    \midrule
    Female & 0.084\\
    Male & 0.316\\

    \bottomrule
  \end{tabular}
  
  \begin{tabular}{llllll}
    \toprule
    \multicolumn{2}{c}{True Values} \\
    \cmidrule(r){1-2}
    Model & TPR & TNR & FPR & FNR \\
    \midrule

    LR & 0.446 & 0.942 & 0.058 & 0.554 \\
    RF & 0.523 & 0.906 & 0.094 & 0.477 \\

    \cmidrule(r){1-2}
    Population & Prior\\
    \midrule
    Female & 0.110\\
    Male & 0.305\\
    \bottomrule
  \end{tabular}
  \vspace{5mm}
   \caption{Top are the estimated parameters (TPR, TNR, prior) for both models. Below are the known values from the training environment in the Adult data set.}
   \label{table1_adult}
\end{table}

\begin{table}[!htb]
  \label{table3_adult}
  \centering
  \begin{tabular}{llllll}
    \toprule
    Parameter & CI & Mean & SD \\
    \midrule

    $\hat{\theta}_1$ & (0.0656, 0.103) & 0.084 & 0.010\\ 
    $\hat{\theta}_2$  & (0.293, 0.338), & 0.316 & 0.011 \\
    $\hat{\alpha}_1$ & (1.09e-4, 0.015) & 0.008 & 0.004 \\ 
    $\hat{\alpha}_2$  & (5.15e-6, 0.020) & 0.010 & 0.006 \\ 
    $\hat{\beta}_1$  & (0.353, 0.428) & 0.392 & 0.019 \\ 
    $\hat{\beta}_2$  & (0.165, 0.247) & 0.207 & 0.021 \\
    \bottomrule
  \end{tabular}
  \vspace{5mm}
  \caption{The estimated parameters and standard deviations from the Hui-Walter paradigm on static data for the Adult data set.}
\end{table}

The Hui-Walter method obtained close estimates of some of the parameters.  The $95\%$ confidence intervals for the rates of adults with over $\$50\text{k}$ income contained the ground truth rate for the male population (${\theta}_2$) and was 0.007 within the ground truth rate of the female population (${\theta}_1$).  The confidence intervals for the error rate estimates ($\hat{\alpha}_1$, $\hat{\alpha}_2$, $\hat{\beta}_1$, $\hat{\beta}_2$) did not contain the actual values.  However, $\hat{\alpha}_1$ and $\hat{\alpha}_2$ underestimated the actual by 7-10 times (0.008 compared to 0.058; 0.010 compared to 0.094) but $\hat{\beta}_1$ and $\hat{\beta}_2$ were closer, underestimating the actual by 2 times (0.392 compared to 0.554; 0.207 compared to 0.477).

\subsection{Summary of Experimental Results}
Our experiments show that the model's false positive and false negative rates and prior can be estimated on unknown data sets with the Hui-Walter method.  Although we assumed the default distribution $Beta (1,1)$ for the errors and priors, the results from the two experiments are generally promising for rough order of magnitude estimates. With more iterations or a more specific distributional assumption, the results would improve.  The results would also likely be more favorable with better trained models as well.


\section{Hui-Walter Online}

\begin{figure}[!htb]
  \centering
  \includegraphics[scale=0.4]{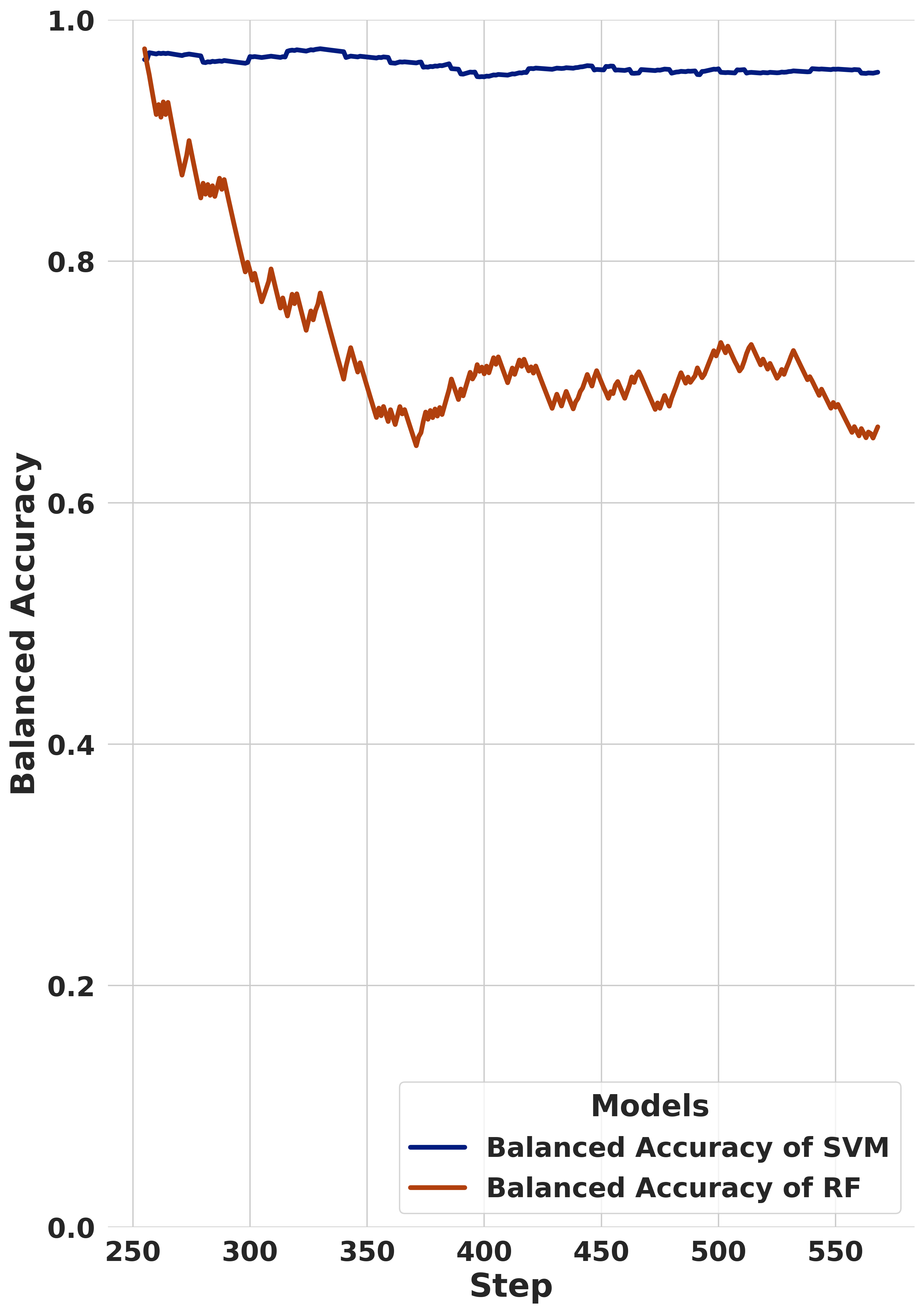}
  \caption{Online-Hui Walter applied to the Wisconsin Breast Cancer data set in a stream with the \texttt{river} python library for online machine learning \cite{river} to simulate online data and the balanced accuracy for both classifiers was calculated over two latent subpopulations starting after a burn in period of $250$ streaming samples. }
  \label{fig4}
\end{figure}

Our main goal in this research is to answer how we estimate false positive and false negative rates of binary categorizers to a stream of unlabeled data without available ground truth. Current online machine learning theory still assumes that labels are available for the data in question \cite{onlineML}. Figure \ref{fig4} shows an experiment where we streamed the Wisconsin Breast Cancer Data set with the \texttt{river} python library for online machine learning \cite{river} \footnote{https://github.com/online-ml/river}.

These features are different from the features used in the non-streaming experiment. The support vector machine is $90\%$ accurate with $\alpha_1 = FPR = 0.18$ and $\beta_2 = FNR = 0.06$. The Random Forest is $88\%$ accurate with $\alpha_2 = FPR = 0.23$ and $\beta_2 = FNR = 0.09$. 

The categorizers then applied predictions to the online data and, for each time step, the contingency table was updated, the new sample was recorded, and latent profile analysis was applied to the new sample, plus the complete history of the samples. Because latent profiles are calculated when a new sample arrives, the latent class of a single sample will often flip. Due to this, Figure \ref{fig4} shows the history of the parameters of interest calculated for the last latent profile. Additionally, due to the instability of the discriminant for Hui-Walter, the first $250$ time steps are omitted because they contain discontinuities. The literature on Hui-Walter states that Hui-Walter may be better at estimating the test agreement than the actual false positive and false negative rates \cite{agreement}, and the results of our experiment in Figure \ref{fig4} show the parameters that come close to the desired metrics of the training environment scaled up or down by a factor of $2$ for $\alpha_i, \beta_i$ for $i=1,2$. Prior probabilities $\theta_1, \theta_2$ are more stable. Empirically, the priors are $\theta_1 = 0.11$ and $\theta_2=0.51$, calculated by taking the number of true values from the labels available in our experimental setup.

\noindent
Table \ref{mae} contains the mean absolute error of the values estimated with Hui-Walter versus those estimated in the training environment. We calculate the mean absolute error using the last $200$ steps. This technique is similar to the burn-in period in Bayesian statistics
\footnote{Data and Experiments available: https://github.com/kslote1/hui-walter}. 

The Rand index (RI) is a measure of the similarity between two cluster assignments, typically the ground truth labels and the clustering results \cite{Rand1971}. It is calculated by considering the number of pairs of data points that are either in the same cluster in both assignments (true positives, TP) or in different clusters in both assignments (true negatives, TN). Given the total number of pairs of data points, N, the Rand index can be computed as follows.

$$RI = \frac{TP + TN}{\binom{N}{2}}$$

The RI ranges from 0 to 1, where a value of 1 indicates perfect agreement between the two cluster assignments, and a value of 0 suggests that the assignments are completely dissimilar.

Now, let us discuss the relationship between the Rand index and accuracy. In the context of binary classification, precision is the proportion of true positive (TP) and true negative (TN) instances out of the total number of instances (P). It can be calculated as:

$$\text{Balanced Accuracy}= \frac{TPR + TNR}{2}$$

In clustering problems, the Rand index can be considered a measure of the ``accuracy" of a clustering algorithm when compared to the ground truth labels. However, it is essential to note that there are some differences between these two measures.

The Rand index is designed for clustering problems and considers both true positive and true negative pairs, whereas the accuracy is designed for classification problems and is based on the number of correctly classified instances.
The Rand index is a measure of similarity between two cluster assignments and does not directly evaluate the quality of a single clustering result. On the contrary, accuracy directly evaluates the performance of a classification model.
In summary, while the Rand index shares some similarities with accuracy, they are designed for different types of problem (clustering versus classification), and their interpretations and use cases are not entirely the same.

\begin{table}[!htb]
\centering
\begin{tabular}{lccc}
\hline
Classifier & \hspace{1mm} Balanced Accuracy  & \hspace{1mm} Rand Index & \hspace{1mm} Accuracy \\ \hline
RF & 0.96 & 0.78 & 0.86 \\
SVM & 0.73 & 0.82 & 0.90 \\ \hline
\end{tabular}
\vspace{5mm}
\caption{Wisconsin Breast Cancer Data Set: Comparison of the mean Balanced Accuracy estimated by Hui-Walter on the online data, Rand Index, and Ground Truth Accuracy for Random Forest and Support Vector Machine.}
\label{table:comparison_wbcd}
\end{table}

\begin{table}[!htb]
\centering
\begin{tabular}{lccc}
\hline
Classifier & \hspace{1mm} Balanced Accuracy  & \hspace{1mm} Rand Index & \hspace{1mm} Accuracy \\ \hline
LR & 0.76 & 0.71 & 0.82 \\
RF & 0.77 & 0.70 & 0.81 \\ \hline
\end{tabular}
\vspace{5mm}
\caption{Adult Data Set: Comparison of the mean Balanced Accuracy estimated by Hui-Walter on the online data, Rand Index, and Ground Truth Accuracy for Logistic Regression and Random Forest}
\label{table:comparison_adult}
\end{table}

\noindent
In order to compare our method to established methodologies from unsupervised learning, we compare the balanced accuracy found for our online implementation of Hui-Walter with the Rand Index which has been shown to relate the accuracy to the unsupervised learning setting \cite{supervising_unsupervised_learning}.  For the Wisconsin Breast Cancer data set, we compare the balanced accuracy for the support vector machine and the random forest to the Rand index for both classifiers over both latent sub-populations. For the random forest model, we found a Rand index of 0.78 and a balanced accuracy of 0.96 from Hui-Walter, and the ground truth accuracy is 0.86. For the support vector machine, the balanced accuracy is 0.73, the RI is 0.82, and the ground truth accuracy is 0.90.  These can be seen in Table \ref{table:comparison_wbcd} where we can see that online Hui-Walter gives a clear improvement over perhaps the only available baseline metric for this problem.  The results are even more compelling for the analogous experiment run on the adult data set; as we can see in Table \ref{table:comparison_adult}, the Balanced Accuracy is closer to the ground truth accuracy than the Rand index.  The classifiers in question (Random Forest, Support Vector Machine, and Logistic Regression) were all chosen for their simplicity and ubiquity in production settings.

\begin{table}[H]
\centering
\label{mae}
\begin{tabular}{|l|l|}
\hline
Parameter  & MAE  \\ \hline
$\theta_1$ & 0.35 \\ \hline
$\theta_2$ & 0.03 \\ \hline
$\alpha_1$ & 0.10 \\ \hline
$\alpha_2$ & 0.15 \\ \hline
$\beta_1$  & 0.05 \\ \hline
$\beta_2$  & 0.09 \\ \hline
\end{tabular}
\vspace{5mm}
\caption{After streaming the data, the parameters that the model ended on compared to the known parameter values through Mean Absolute Error.}
\end{table}

Although this test yielded reasonable estimates for most parameters, the main issue with this approach for online estimation is the need for more standard errors, which is given by Bayesian estimation for the static case \cite{Johnson2001}. The worst estimate on the streaming data, $\theta_1$, also closely matches the static data case found with the Gibbs sampling method.

\section{Conclusions}
\label{sec:conclusions}

Our results show that the Hui-Walter method works very well for static data and gives \textit{plausible} results for online data, and this method requires further improvements. The data from this experiment are in a three-way contingency table, and a log-linear model could improve the estimate as is in \cite{hui2}. Additionally, the algebraic geometry of the three-way contingency tables could yield different results, as there is a relationship between tensor factorization and the parameters of the product-multinomial model \cite{TensorDunson2009}. One of the limitations of using explicit formulas for the $2\times 2 \times 2$ case is that sometimes the solutions do not have \textit{plausible} solutions \cite{Hui1980}. In other words, it is possible to get a solution for the false positive rate greater than one or less than zero. Furthermore, the performance of the straightforward solutions on the online data lacked confidence intervals, which are problematic for applied settings, and the solutions are not as precise as the Gibbs sampler. One suggestion for further research on how this can be improved is to leverage online Gibbs sampling \cite{onlineGIBBS,OnlineGibbsDupuy2017}.

We have shown how to use the Hui-Walter paradigm to estimate false positive and false negative rates and prior probabilities when no ground truth is available. One of the core assumptions of statistical reasoning and its offshoot machine learning is that all of the models' possible data is collected ahead of time, and a practitioner only needs to sample appropriate training and holdout test sets. This assumption is far from the reality of many machine learning applications worldwide. Machine learning practitioners increasingly adapt to the reality that models make predictions based on previously unseen unlabeled data that change in the shape of the distribution over time. In this paper, we tackle this very common problem in applied machine learning. Specifically, we derived a way to measure the efficacy or effectiveness of machine learning models in a situation where the data set is unknown and there are no assumptions about the data distribution. We have also shown how this methodology applies to static and streaming data.


\section{Limitations}

The Hui-Walter method and the online methods mentioned in this text use data sets with even distributions of classes. Further work should be and considerations should be made on heavily unbalanced classes or classes where the prevalence of the phenomenon is very low.

For the Wisconsin Breast Cancer Data Set, it is essential to note that the data set has some limitations. For example, data are based on a specific population of patients and may not represent all cases of breast cancer. Furthermore, the data are from the 1990s and may not reflect more recent advances in breast cancer diagnosis and treatment. Despite these limitations, the Wisconsin breast cancer data set remains an essential resource for researchers studying breast cancer and working to improve the diagnosis and treatment of this disease. It has contributed significantly to our understanding of breast cancer and will continue to be a valuable resource for researchers in the field.

\section*{Acknowledgments}
We would like to acknowledge Yichuan Zhao, Herbert Roitblatt, and Igor Belykh for their valuable feedback.

\bibliographystyle{splncs04}

\bibliography{references}

\begin{thebibliography}{10}
\providecommand{\url}[1]{\texttt{#1}}
\providecommand{\urlprefix}{URL }
\providecommand{\doi}[1]{https://doi.org/#1}

\bibitem{online_linear_regression}
Abu-Shaira, M., Speegle, G.: Online linear regression based on weighted average. In: Han, H., Baker, E. (eds.) Next Generation Data Science. pp. 88--108. Springer Nature Switzerland (Jun 2024)

\bibitem{benarous2024online}
Arous, G.B., Gheissari, R., Jagannath, A.: Online stochastic gradient descent on non-convex losses from high-dimensional inference. Journal of Machine Learning Research  \textbf{22},  1--52 (Jan 2024)

\bibitem{awe}
Banfield, J.D., Raftery, A.E.: Model-based {G}aussian and non-{G}aussian clustering. Biometrics  \textbf{49},  803--821 (Sep 1993). \doi{10.2307/2532201}

\bibitem{agreement}
Bertrand, P., Bénichou, J., Grenier, P., Chastang, C.: Hui and {W}alter's latent-class reference-free approach may be more useful in assessing agreement than diagnostic performance. Journal of Clinical Epidemiology  \textbf{58},  688--700 (Jul 2005). \doi{10.1016/j.jclinepi.2004.10.021}

\bibitem{Blasques2018}
Blasques, F., Gorgi, P., Koopman, S.J., Wintenberger, O.: Feasible invertibility conditions and maximum likelihood estimation for observation-driven models. Electronic Journal of Statistics  \textbf{12},  1019--1052 (Jan 2018). \doi{10.1214/18-EJS1416}

\bibitem{mulit_colin}
Chan, J.Y.L., Leow, S.M.H., Bea, K.T., Cheng, W.K., Phoong, S.W., Hong, Z.W., Chen, Y.L.: Mitigating the multicollinearity problem and its machine learning approach: A review. Mathematics  \textbf{10}(8) (Apr 2022). \doi{10.3390/math10081283}

\bibitem{pmlr_online_k_means}
Cohen-Addad, V., Guedj, B., Kanade, V., Rom, G.: Online k-means clustering. In: Banerjee, A., Fukumizu, K. (eds.) Proceedings of The 24th International Conference on Artificial Intelligence and Statistics. Proceedings of Machine Learning Research, vol.~130, pp. 1126--1134. PMLR (Apr 2021)

\bibitem{Davis2020}
Davis, J.: Posterior adaptation with new priors. arXiv preprint arXiv:2007.01386  (Jul 2020), \url{http://arxiv.org/abs/2007.01386}

\bibitem{uci}
Dua, D., Graff, C.: {UCI} machine learning repository (2019). \doi{10.24432/C5DW2B}, \url{https://archive.ics.uci.edu/ml}

\bibitem{TensorDunson2009}
Dunson, D.B., Xing, C.: Nonparametric {B}ayes modeling of multivariate categorical data. Journal of the American Statistical Association  \textbf{104},  1042--1051 (Jan 2009). \doi{10.1198/jasa.2009.tm08439}

\bibitem{OnlineGibbsDupuy2017}
Dupuy, C., Bach, F.: Online but accurate inference for latent variable models with local gibbs sampling. Journal of Machine Learning Research  \textbf{18},  1--45 (Nov 2017), \url{http://jmlr.org/papers/v18/16-374.html}

\bibitem{enola2000}
Enøe, C., Georgiadis, M.P., Johnson, W.O.: Estimation of sensitivity and specificity of diagnostic tests and disease prevalence when the true disease state is unknown. Preventive Veterinary Medicine  \textbf{45},  61--81 (May 2000)

\bibitem{cophenetic}
Farris, J.S.: On the cophenetic correlation coefficient. Systematic Biology  \textbf{18}(3),  279--285 (Sep 1969). \doi{10.2307/2412324}

\bibitem{supervising_unsupervised_learning}
Garg, V., Kalai, A.T.: Supervising unsupervised learning. In: Advances in Neural Information Processing Systems. NIPS'18, vol.~31, pp. 4996--5006. Curran Associates, Inc., Red Hook, NY, USA (Dec 2018)

\bibitem{gibbs_marginals}
Gelfand, A.E., Smith, A.F.M.: Sampling-based approaches to calculating marginal densities. Journal of the American Statistical Association  \textbf{85}(410),  398--409 (Jun 1990), \url{https://www.jstor.org/stable/2289776}

\bibitem{gibbs}
Geman, S., Geman, D.: Stochastic relaxation, gibbs distributions, and the {B}ayesian restoration of images. IEEE Transactions on Pattern Analysis and Machine Intelligence  \textbf{6}(6),  564--584 (Nov 1984). \doi{10.1109/TPAMI.1984.4767596}

\bibitem{onlineML}
Gomes, H.M., Read, J., Bifet, A., Barddal, J.P., Gama, J.: Machine learning for streaming data: State of the art, challenges, and opportunities. SIGKDD Explor. Newsl.  \textbf{21}(2),  6--22 (Nov 2019). \doi{10.1145/3373464.3373470}

\bibitem{burnin}
Hamra, G., MacLehose, R., Richardson, D.: Markov chain monte carlo: an introduction for epidemiologists. International Journal of Epidemiology  \textbf{42}(2),  627--634 (Apr 2013). \doi{10.1093/ije/dyt043}

\bibitem{Hui1980}
Hui, S.L., Walter, S.D.: Estimating the error rates of diagnostic tests. Biometrics  \textbf{36},  167--171 (Mar 1980)

\bibitem{hui2}
Hui, S.L., Zhou, X.H.: Evaluation of diagnostic tests without gold standards. Statistical Methods in Medical Research  \textbf{7},  354--370 (Dec 1998). \doi{10.1177/096228029800700404}

\bibitem{heirarchal_feautre_selection}
Ienco, D., Meo, R.: Exploration and reduction of the feature space by hierarchical clustering. In: Proceedings of the 2008 SIAM International Conference on Data Mining. pp. 577--587. SIAM (Apr 2008). \doi{10.1137/1.9781611972788.53}

\bibitem{Johnson2001}
Johnson, W.O., Gastwirth, J.L., Pearson, L.M.: Screening without a "{G}old {S}tandard": The {H}ui-{W}alter paradigm revisited. American Journal of Epidemiology  \textbf{153},  921--924 (May 2001). \doi{10.1093/aje/153.9.921}

\bibitem{onlineGIBBS}
Kim, Y., Chae, M., Jeong, K., Kang, B., Chung, H.: An online gibbs sampler algorithm for hierarchical dirichlet processes prior. In: Frasconi, P., Landwehr, N., Manco, G., Vreeken, J. (eds.) Machine Learning and Knowledge Discovery in Databases. vol.~9851, pp. 509--523. Springer International Publishing (Sep 2016)

\bibitem{algebra}
Krampe, A., Kuhnt, S.: Model selection for contingency tables with algebraic statistics. In: Gibilisco, P., Riccomagno, E., Rogantin, M.P., Wynn, H.P. (eds.) Algebraic and Geometric Methods in Statistics, pp. 83--97. Cambridge University Press (May 2009). \doi{10.1017/CBO9780511642401}

\bibitem{distance_clustering}
Lu, X.: Information mandala: Statistical distance matrix with clustering. IEEE Access  \textbf{9},  56563--56577 (Apr 2021). \doi{10.1109/ACCESS.2021.3072237}

\bibitem{tsne}
Maaten, L.V.D., Hinton, G.: Visualizing data using t-sne. Journal of Machine Learning Research  \textbf{9},  2579--2605 (Nov 2008)

\bibitem{river}
Montiel, J., Halford, M., Mastelini, S.M.G.B., Sourty, R., Vaysse, R., Zouitine, A., Gomes, H.M., Read, J., Abdessalem, T., Bifet, A.: River: Machine learning for streaming data in python. Journal of Machine Learning Research  \textbf{21}(60), ~1--6 (2020), \url{http://jmlr.org/papers/v21/20-1387.html}

\bibitem{Estimating_accuracy_from_unlabeled_data}
Platanios, E.A., Dubey, A., Mitchell, T.: Estimating accuracy from unlabeled data: a {B}ayesian approach. In: Proceedings of the 33rd International Conference on Machine Learning. ICML'16, vol.~48, pp. 1416--1425. Proceedings of Machine Learning Research (Jun 2016)

\bibitem{Rand1971}
Rand, W.M.: Objective criteria for the evaluation of clustering methods. Journal of the American Statistical Association  \textbf{66},  846--850 (Dec 1971)

\bibitem{saracli2013comparison}
Saraçli, S., Doğan, N., Doğan, I.: Comparison of hierarchical cluster analysis methods by cophenetic correlation. Journal of Inequalities and Applications  \textbf{2013}(1), ~203 (Apr 2013). \doi{10.1186/1029-242X-2013-203}

\bibitem{online_convex}
Shalev-Shwartz, S.: Online learning and online convex optimization. Found. Trends Mach. Learn.  \textbf{4}(2),  107–194 (Feb 2012). \doi{10.1561/2200000018}

\bibitem{Singal2014}
Singal, A.G., Higgins, P.D., Waljee, A.K.: A primer on effectiveness and efficacy trials. Clinical and Translational Gastroenterology  \textbf{5}(e45), ~1 (Jan 2014). \doi{10.1038/ctg.2013.13}

\bibitem{spearman}
Spearman, C.: The proof and measurement of association between two things. The American Journal of Psychology  \textbf{15}(1),  72--101 (Jan 1904), \url{http://www.jstor.org/stable/1412159}

\bibitem{wald}
Vijaya, Sharma, S., Batra, N.: Comparative study of single linkage, complete linkage, and ward method of agglomerative clustering. In: 2019 International Conference on Machine Learning, Big Data, Cloud and Parallel Computing (COMITCon). pp. 568--573. IEEE (Feb 2019). \doi{10.1109/COMITCon.2019.8862232}

\bibitem{feature_selection_from_clusters}
Yan, X., Nazmi, S., Erol, B.A., Homaifar, A., Gebru, B., Tunstel, E.: An efficient unsupervised feature selection procedure through feature clustering. Pattern Recognition Letters  \textbf{131},  277--284 (Mar 2020). \doi{10.1016/j.patrec.2019.12.022}

\end{thebibliography}
\end{document}